\documentclass[11pt,a4paper]{article}
\usepackage[hyperref]{acl2019}
\usepackage{times}
\usepackage{latexsym}
\usepackage{MnSymbol}
\usepackage{amsmath}
\usepackage[capitalize]{cleveref}
\usepackage{amsfonts}
\usepackage{url}
\usepackage{color,soul}
\usepackage{graphicx,bm}
\usepackage{multirow,multicol}
\usepackage[font=small]{caption}
\usepackage{subcaption}
\usepackage{algorithm}
\usepackage[noend]{algpseudocode}
\usepackage{afterpage}
\usepackage{tabularx}
\usepackage{arydshln}
\usepackage{url}
\usepackage[shortlabels]{enumitem}
\usepackage{bigstrut}
\usepackage{array}
\usepackage{microtype}
\usepackage{pifont}
\usepackage{flushend}
\usepackage{afterpage}
\usepackage{tabularx}
\usepackage{arydshln}

\makeatletter
\newcommand{\thickhline}{%
 \noalign {\ifnum 0=`}\fi \hrule height 1pt
 \futurelet \reserved@a \@xhline
}
\def\hlinewd#1{%
	\noalign{\ifnum0=`}\fi\hrule \@height #1 %
	\futurelet\reserved@a\@xhline}

\newcolumntype{"}{@{\hskip\tabcolsep\vrule width 1pt\hskip\tabcolsep}}
\makeatother

\crefformat{section}{\S#2#1#3} % see manual of cleveref, section 8.2.1
\crefformat{subsection}{\S#2#1#3}
\crefformat{subsubsection}{\S#2#1#3}

\definecolor{mypink1}{rgb}{0.858, 0.188, 0.478}
\definecolor{maroon}{RGB}{220, 20, 60}

\aclfinalcopy % Uncomment this line for the final submission
\def\aclpaperid{926} % Enter the acl Paper ID here

\title{MELD: A Multimodal Multi-Party Dataset\\ for Emotion~Recognition in Conversations}

\author{Soujanya Poria$^{\dagger}$, Devamanyu Hazarika$^{\Phi}$, Navonil Majumder$^{\ddagger}$,\\ \textbf{Gautam Naik$^{\mathparagraph}$, Erik Cambria$^{\mathparagraph}$, Rada Mihalcea$^{\iota}$}\\
		$^{\dagger}$Information Systems Technology and Design, SUTD, Singapore \\ 
		$^{\Phi}$School of Computing, National University of Singapore, Singapore \\
$^{\ddagger}$Centro de Investigaci\'on en Computaci\'on, Instituto Polit\'ecnico Nacional, Mexico\\
$^{\mathparagraph}$Computer Science \& Engineering, Nanyang Technological University, Singapore\\
$^{\iota}$Computer Science \& Engineering, University of Michigan, USA\\
{\tt sporia@sutd.edu.sg}, {\tt hazarika@comp.nus.edu.sg},\\ {\tt navo@nlp.cic.ipn.mx}, {\tt gautam@sentic.net},\\ {\tt cambria@ntu.edu.sg}, {\tt mihalcea@umich.edu} \\}

\date{}
\aclfinalcopy
\begin{document}
\maketitle
\begin{abstract}
 Emotion recognition in conversations (ERC) is a challenging task that has recently gained popularity due to its potential applications. % such as dialogue systems, user behavior understanding, and emotional state tracking. 
 Until now, however, there has been no large-scale multimodal multi-party emotional conversational database containing more than two speakers per dialogue. To address this gap, we propose the \emph{Multimodal EmotionLines Dataset} (MELD), an extension and enhancement of EmotionLines. MELD contains about 13,000 utterances from 1,433 dialogues from the TV-series \textit{Friends}. Each utterance is annotated with emotion and sentiment labels, and encompasses audio, visual, and textual modalities. We propose several strong multimodal baselines and show the importance of contextual and multimodal information for emotion recognition in conversations. The full dataset is available for use at \url{http://affective-meld.github.io}.
 
%  \hl{MELD is superior to its well-known counterparts SEMAINE and IEMOCAP as it consists of multi-party conversations and almost twice the number of utterances.} 
\end{abstract}

\section{Introduction} \label{sec:intro}

\begin{figure*}[t] 
	\centering 
	\includegraphics[scale=0.36]{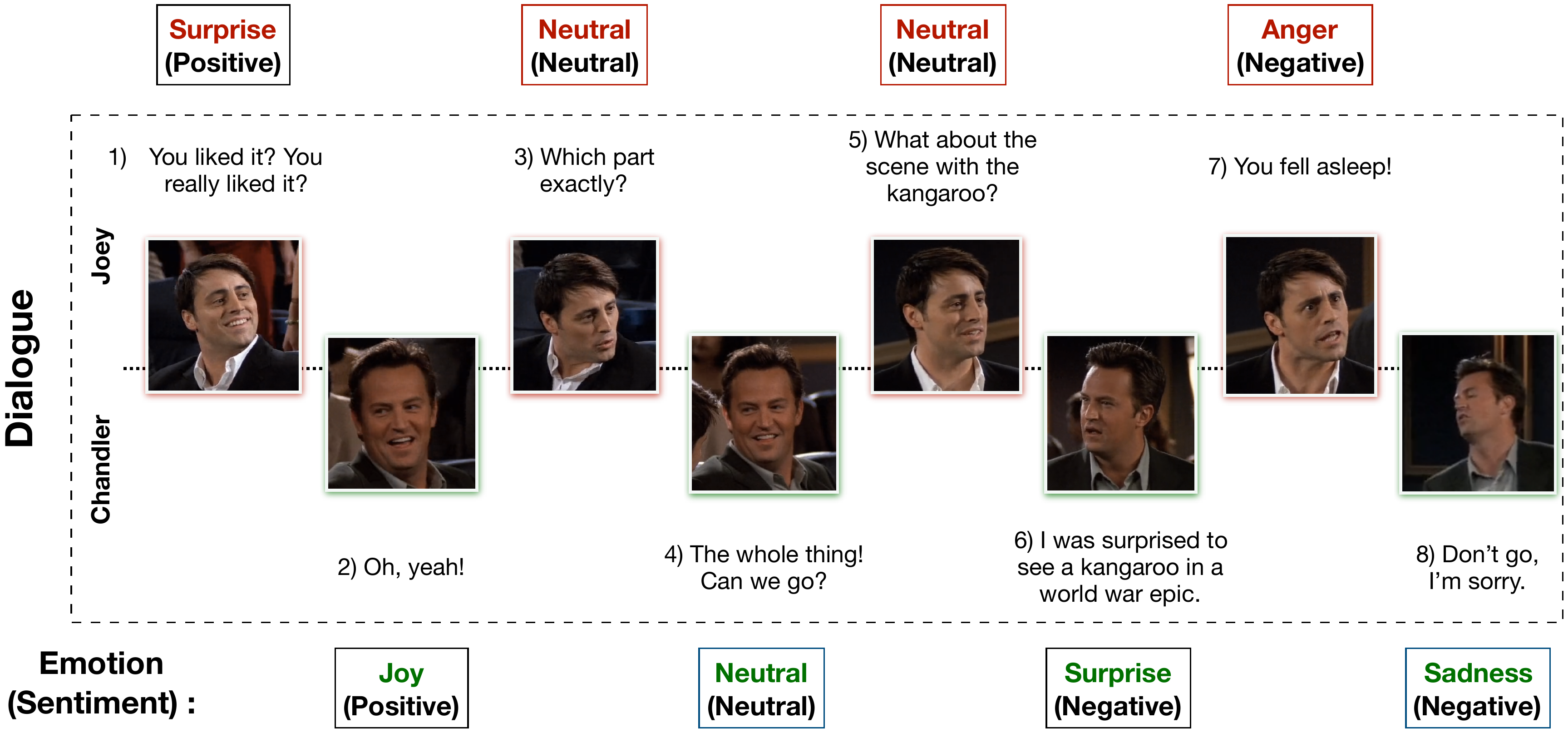} 
	\caption[]{Emotion shift of speakers in a dialogue in comparison with their previous emotions.}
	\label{fig:emoc}
	%\vspace{-0.2cm}
\end{figure*}
\begin{figure}[t] 
	\centering 
	\includegraphics[width=\linewidth]{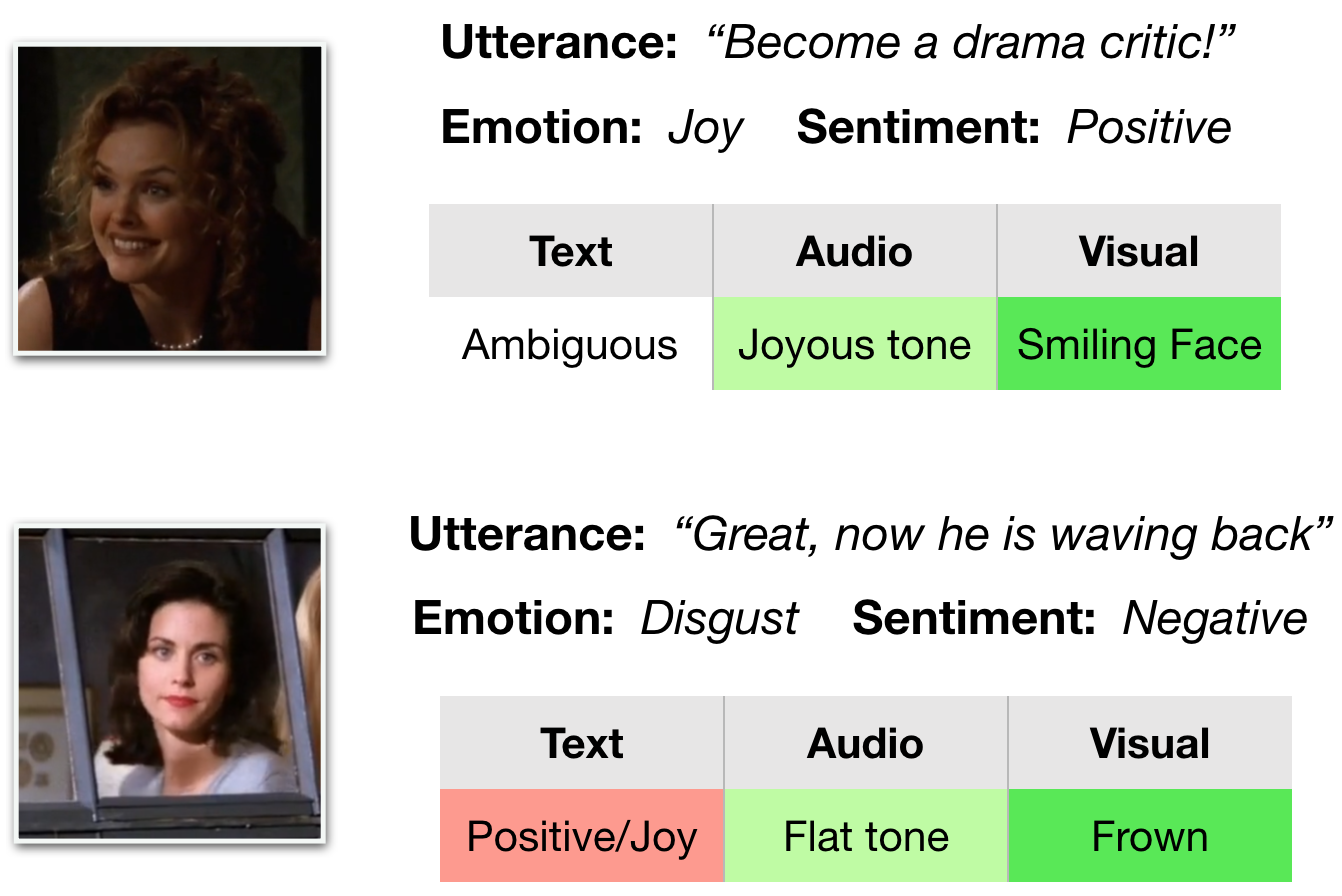} 
	\caption[]{Importance of multimodal cues. Green shows primary modalities responsible for sentiment and emotion.}
	\label{fig:multimodal_examples}
%	\vspace{-0.4cm}
\end{figure}
With the rapid growth of Artificial Intelligence (AI), multimodal emotion recognition has become a major research topic, primarily due to its potential applications in many challenging tasks, such as dialogue generation, user behavior understanding, multimodal interaction, and others. A conversational emotion recognition system can be used to generate appropriate responses by analyzing user emotions~\cite{zhou2017emotional,rashkin2018know}.

Although significant research work has been carried out on multimodal emotion recognition using audio, visual, and text modalities~\citep{zadeh2016deep,wollmer2013youtube}, significantly less work has been devoted to emotion recognition in conversations (ERC). One main reason for this is the lack of a large multimodal conversational dataset. 

According to~\citet{poria2019emotion}, ERC presents several challenges such as conversational context modeling, emotion shift of the interlocutors, and others, which make the task more difficult to address. Recent work proposes solutions based on multimodal memory networks~\cite{hazarika2018conversational}. However, they are mostly limited to dyadic conversations, and thus not scalable to ERC with multiple interlocutors. This calls for a multi-party conversational data resource that can encourage research in this direction.

In a conversation, the participants' utterances generally depend on their conversational context. This is also true for their associated emotions. In other words, the context acts as a set of parameters that may influence a person to speak an utterance while expressing a certain emotion. Modeling this context can be done in different ways, e.g., by using recurrent neural networks (RNNs) and memory networks~\cite{hazarika2018conversational,porcon,serban2017hierarchical}. Figure~\ref{fig:emoc} shows an example where the speakers change their emotions (emotion shifts) as the dialogue develops. The emotional dynamics here depend on both the previous utterances and their associated emotions. For example, the emotion shift in utterance eight (in the figure) is hard to determine unless cues are taken from the facial expressions and the conversational history of both  speakers. Modeling such complex inter-speaker dependencies is one of the major challenges in conversational modeling.

Conversation in its natural form is multimodal. In dialogues, we rely on others' facial expressions, vocal tonality, language, and gestures to anticipate their stance. For emotion recognition, multimodality is particularly important. For the utterances with language that is difficult to understand, we often resort to other modalities, such as prosodic and visual cues, to identify their emotions. Figure~\ref{fig:multimodal_examples} presents examples from the dataset where the presence of multimodal signals in addition to the text itself is necessary in order to make correct predictions of their emotions and sentiments.

Multimodal emotion recognition of sequential turns encounters several other challenges. One such example is the classification of short utterances. Utterances like ``\textit{yeah}'', ``\textit{okay}'', ``\textit{no}'' can express varied emotions depending on the context and discourse of the dialogue. However, due to the difficulty of perceiving emotions from text alone, most models resort to assigning the majority class (e.g., \textit{non-neutral} in EmotionLines). Approximately $42\%$ of the utterances in MELD are shorter than five words. We thus provide access to the multimodal data sources for each dialogue and posit that this additional information would benefit the emotion recognition task by improving the context representation and supplementing the missing or misleading signals from other modalities. Surplus information from attributes such as the speaker's facial expressions or intonation in speech could guide models for better classification. We also provide evidence for these claims through our experiments.

The development of conversational AI thus depends on the use of both contextual and multimodal information. The publicly available datasets for multimodal emotion recognition in conversations -- IEMOCAP and SEMAINE -- have facilitated a significant number of research projects, but also have limitations due to their relatively small number of total utterances and the  lack of multi-party conversations. There are also other multimodal emotion and sentiment analysis datasets, such as MOSEI~\cite{zadeh2018multimodal}, MOSI~\cite{zadeh2016multimodal}, and MOUD~\cite{perez2013utterance}, but they contain individual narratives instead of dialogues. On the other hand, EmotionLines~\citep{chen2018emotionlines} is a dataset that contains dialogues from the popular TV-series \textit{Friends} with more than two speakers. However, EmotionLines can only be used for textual analysis as it does not provide data from other modalities.

In this work, we extend, improve, and further develop the EmotionLines dataset for the multimodal scenario. We propose the \textit{Multimodal EmotionLines Dataset} (MELD), which  includes not only textual dialogues, but also their corresponding visual and audio counterparts. This paper makes several contributions:
\begin{itemize}[leftmargin=*]
\itemsep0em 
\item MELD contains multi-party conversations that are more challenging to classify than dyadic variants available in previous datasets. % IEMOCAP and SEMAINE.
\item There are more than 13,000 utterances in MELD, which makes our dataset nearly double the size of  existing multimodal conversational datasets.
\item MELD provides multimodal sources and can be used in a multimodal affective dialogue system for enhanced grounded learning.
\item We establish a strong baseline, proposed by \citet{majumder2018dialoguernn}, which is capable of emotion recognition in multi-party dialogues by inter-party dependency modeling.
\end{itemize}

The remainder of the paper is organized as follows: Section~\ref{sec:emotionlines} illustrates the EmotionLines dataset; we then present MELD in Section~\ref{sec:meld}; strong baselines and experiments are elaborated in Section~\ref{sec:experiment}; future directions and applications of MELD are covered in Section~\ref{sec:future} and~\ref{sec:appli}, respectively; finally, Section~\ref{sec:conclusion} concludes the paper.
% Multimodal data analysis exploits information from multiple-parallel data channels for decision making.

% The emotion change and emotion flow in the sequence of turns in a dialogue can make context modeling difficult. 

% \citet{hazarika2018conversational} claimed that it is not enough to just use an LSTM or any other network that takes all the previous utterances as input and generates a vector to represent the context. According to them, a conversational model should know the speaker of each utterance and they experimentally showed that this helps in producing better context representation relevant to emotion recognition by means of inter-speaker dependency modeling. Their model dynamically attends to the history of utterances by the same speaker or the other speaker for emotion recognition.

\section{EmotionLines Dataset}
\label{sec:emotionlines}

The MELD dataset has evolved from the EmotionLines dataset developed by \citet{chen2018emotionlines}. EmotionLines contains dialogues from the popular sitcom \textit{Friends}, where each dialogue contains utterances from multiple speakers. 

EmotionLines was created by crawling the dialogues from each episode and then grouping them based on the number of utterances in a dialogue into four groups of [5, 9], [10, 14], [15, 19], and [20, 24] utterances respectively. Finally, 250 dialogues were sampled randomly from each of these groups, resulting in the final dataset of 1,000 dialogues.

\subsection{Annotation}
The utterances in each dialogue were annotated with the most appropriate emotion category. For this purpose, Ekman's six universal emotions (\emph{Joy, Sadness, Fear, Anger, Surprise, and Disgust}) were considered as annotation labels. This annotation list was extended with two additional emotion labels: \emph{Neutral} and \textit{Non-Neutral}. 

Each utterance was annotated by five workers from the Amazon Mechanical Turk (AMT) platform. A majority voting scheme was applied to select a final emotion label for each utterance. The overall Fleiss' kappa score of this annotation process was $0.34$.

\begin{table}[h]
	\centering
		\resizebox{\linewidth}{!}{
	\begin{tabular}{c|c|c|c|c|c|c}
		\hlinewd{1.5pt}
			\multirow{2}{*}{Dataset}&\multicolumn{3}{c|}{$\#$ Dialogues}&\multicolumn{3}{c}{$\#$ Utterances}\\
			&train&dev&test&train&dev&test\\
      \hlinewd{0.8pt}
		  EmotionLines &720&80&200&10561&1178&2764\\
		  MELD     &1039&114&280&9989&1109&2610\\
		\hlinewd{0.8pt}
	\end{tabular}
	}
	\caption{Comparison between the original EmotionLines dataset and MELD.}
	\label{tab:data}
\end{table}

\section{Multimodal EmotionLines Dataset (MELD)}
\label{sec:meld}

We start the construction of the MELD corpus by extracting the starting and ending timestamps of all utterances from every dialogue in the EmotionLines dataset. To accomplish this, we crawl through the subtitles of all the episodes and heuristically extract the respective timestamps. In particular, we enforce the following constraints:
	\begin{enumerate}[leftmargin=*]
  \itemsep0em 
		\item Timestamps of the utterances in a dialogue must be in an increasing order.
		\item All the utterances in a dialogue have to belong to the same episode and scene.
	\end{enumerate}

These constraints revealed a few outliers in EmotionLines where some dialogues span across scenes or episodes. For example, the dialogue in Table~\ref{tab:errorEmo} contains two natural dialogues from episode 4 and 20 of season 6 and 5, respectively. We decided to filter out these anomalies, thus resulting in a different number of total dialogues in MELD as compared to EmotionLines (see Table~\ref{tab:data}).
  
Next, we employ three annotators to label each utterance, followed by a majority voting to decide the final label of the utterances. We drop a few utterances where all three annotations were different, and also remove their corresponding dialogues to maintain coherence. A total of 89 utterances spanning 11 dialogues fell under this category.
	
Finally, after obtaining the timestamp of each utterance, we extract their corresponding audio-visual clips from the source episode followed by the extraction of audio content from these clips. We format the audio files as 16-bit PCM WAV files for further processing. The final dataset includes visual, audio, and textual modalities for each utterance.\footnote{We consulted a legal office to verify that the usage and distribution of very short length videos fall under the \textit{fair use} category.}

\begin{table*}[h]
	\centering
	\small
	\resizebox{\linewidth}{!}{
	\begin{tabular}{ccccc}
		\hlinewd{1.5pt}
		Episode&Utterance& Speaker& Emotion& Sentiment \\
		\hlinewd{0.8pt}
% 		\parbox[t]{2mm}{\multirow{6}{*}{\rotatebox[origin=c]{90}{S6.E4}}}&Hey Estelle, listen & Joey & neutral & neutral \\
% 		&Well! Well! Well! Joey Tribbiani! So you came back huh? They & Estelle & surprise & positive\\
		\parbox[c]{2mm}{\multirow{6}{*}{\rotatebox[origin=c]{90}{S6.E4}}}&What are you talkin’ about? I never left you! You’ve always been my agent! & Joey & surprise & negative \\
		&Really?! & Estelle & surprise & positive\\
		&Yeah! & Joey & joy & positive\\
		&Oh well, no harm, no foul. & Estelle & neutral & neutral\\ %\cdashline{2-5}
		\hdashline
		\parbox[t]{2mm}{\multirow{3}{*}{\rotatebox[origin=c]{90}{S5.E20}}}&\textcolor{maroon}{Okay, you guys free tonight?} & \textcolor{maroon}{Gary} & \textcolor{maroon}{neutral}&\textcolor{maroon}{neutral} \\
		&\textcolor{maroon}{Yeah!!} & \textcolor{maroon}{Ross} & \textcolor{maroon}{joy} & \textcolor{maroon}{positive}\\
		&\textcolor{maroon}{Tonight? You-you didn't say it was going to be at nighttime.} & \textcolor{maroon}{Chandler} & \textcolor{maroon}{surprise}& \textcolor{maroon}{negative}\\
		\hlinewd{.8pt}
	\end{tabular}
}
	\caption{A dialogue in EmotionLines where utterances from two different episodes are present. The first four utterances in this dialogue have been taken from episode 4 of season 6. The last three utterances in red font are from episode 20 of season 5.}
	\label{tab:errorEmo}
\end{table*}

\subsection{Dataset Re-annotation}

The utterances in the original EmotionLines dataset were annotated by looking only at the transcripts. However, due to our focus on multimodality, we re-annotate all the utterances by asking the three annotators to also look at the available video clip of the utterances. We then use  majority-voting  to obtain the final label for each utterance.

The annotators were graduate students with high proficiency in English speaking and writing. Before starting the annotation, they were briefed about the annotation process with a few examples. 

We achieve an overall Fleiss' kappa score of $0.43$ which is higher than the original EmotionLines annotation whose kappa score was $0.34$ (kappa of IEMOCAP annotation process was $0.4$), thus suggesting the usefulness of the additional modalities during the annotation process. 

2,772 utterances in the EmotionLines dataset were labeled as \emph{non-neutral} where the annotators agreed that the emotion is not neutral but they could not reach agreement regarding the correct emotion label. This hampers classification, as the \emph{non-neutral} utterance space and the other emotion-label spaces get conflated. In our case, we remove the utterances where the annotators fail to reach an agreement on the definite emotion label. 

The number of disagreements in our annotation process is 89, which is much lower than the 2,772 disagreements in EmotionLines, reflecting again the annotation improvement obtained through a multimodal dataset. Table~\ref{tab:annotation} shows examples of utterances where the annotators failed to reach consensus.

\begin{table}[b]
	\small
	\resizebox{\linewidth}{!}{
	\begin{tabular}{cccc}
		\hlinewd{1.5pt}
		Utterance & Annotator 1 & Annotator 2 & Annotator 3\\
		\hlinewd{0.8pt}
		You know? Forget it! & sadness & disgust & anger \\ 
	  Oh no-no, give me & \multirow{2}{*}{anger} & \multirow{2}{*}{sadness} & \multirow{2}{*}{neutral} \\
		some specifics. & & & \\ 
		I was surprised to see a & \multirow{2}{*}{surprise} & \multirow{2}{*}{anger} & \multirow{2}{*}{joy}\\
		kangaroo in a World War epic. & & &\\ 
		Or, call an ambulance. & anger & surprise & neutral\\ 
		\hlinewd{0.8pt}
	\end{tabular}
}
	\caption{Some examples of the utterances for which annotators could not reach consensus.}
	\label{tab:annotation}
\end{table}

Table~\ref{tab:emomeldvsemotionlines} shows the label-wise comparison between EmotionLines and MELD dataset. For most of the utterances in MELD, the annotations match the  original annotations in EmotionLines. Yet, there exists a significant amount of samples whose utterances have been changed in the re-annotation process. For example, the utterance \emph{This guy fell asleep!} (see Table~\ref{tab:emovsmeld}), was labeled as \emph{non-neutral} in EmotionLines but after viewing the associated video clip, it is correctly re-labeled as \emph{anger} in MELD. 

The video of this utterance reveals an angry and frustrated facial expression along with a high vocal pitch, thus helping to recognize its correct emotion. The annotators of EmotionLines had access to the context, but this was not sufficient, as the availability of additional modalities can sometime bring more information for the classification of such instances. %Also, the annotators for MELD provided feedback that having access to the video clips during the annotation process indeed helps. 
These scenarios justify both \textit{context} and \textit{multimodality} to be important aspects for emotion recognition in conversation.

\begin{table}[t]
	\centering
	\small
	\resizebox{\linewidth}{!}{
	\begin{tabular}{cc|ccc|ccc}
		\hlinewd{1.5pt}
		&& \multicolumn{3}{c}{{EmotionLines}} & \multicolumn{3}{c}{{MELD}} \\
		\cline{3-8}
		\multicolumn{2}{c|}{{Categories}} & Train & Dev & Test & Train & Dev & Test \\
    \hlinewd{0.8pt}
  	\parbox[t]{2mm}{\multirow{7}{*}{\rotatebox[origin=c]{90}{\scriptsize{Emotion}}}}&anger & 524  & 85 & 163 & 1109  & 153 & 345 \\
		&disgust & 244  & 26 & 68 & 271  & 22 & 68  \\
		&fear  & 190  & 29 & 36  & 268  & 40 & 50  \\
		&joy   & 1283  & 123 & 304  & 1743  & 163 & 402  \\
		&neutral & 4752  & 491 & 1287  & 4710  & 470 & 1256 \\
		&sadness & 351  & 62 & 85  & 683  & 111 & 208 \\
		&surprise & 1221  & 151 & 286  & 1205  & 150 & 281 \\
	%	non-neutral & 2017 & 214 & 541
		\hlinewd{0.8pt}
		\parbox[t]{2mm}{\multirow{3}{*}{\rotatebox[origin=c]{90}{\scriptsize{Sentiment}}}}&negative&-&-&-& 2945 & 406 & 833 \\
		& neutral &-&-&-& 4710 & 470 & 1256 \\
		&positive &-&-&-& 2334 & 233 & 521 \\
		\hlinewd{0.8pt}
	\end{tabular}
	}
	\caption{Emotion and Sentiment distribution in MELD vs. EmotionLines.}
	\label{tab:emomeldvsemotionlines}
\end{table}
\paragraph{Timestamp alignment.}
There are many utterances in the subtitles that are grouped within identical timestamps in the subtitle files. In order to find the accurate timestamp for each utterance, we use a transcription alignment tool \emph{Gentle},\footnote{\scriptsize{\url{http://github.com/lowerquality/gentle}}} which automatically aligns a transcript with the audio by extracting word-level timestamps from the audio (see Table~\ref{tab:pruning}). In \cref{tab:dataformat},
we show the final format of the MELD dataset.

\paragraph{Dyadic MELD.} We also provide another version of MELD where all the non-extendable contiguous dyadic sub-dialogues of MELD are extracted. For example, let a three-party dialogue in MELD with speaker ids $1,2,3$ have their turns in the following order: $[1,2,1,2,3,2,1,2]$. 

From this dialogue sequence, dyadic MELD will have the following sub-dialogues as samples: $[1,2,1,2], [2,3,2]$ and $[2,1,2]$. %However, it should be noted that the baseline results on this dataset have been obtained using only the multiparty variant of MELD.
However, the reported results in this paper are obtained using only the multiparty variant of MELD.

% For example, a dialogue in MELD with three speakers, speaking in the sequence $[1,2,1,2,3,2,1,2]$, has dyadic these sub-dialogues in dyadic MELD --- $[1,2,1,2], [2,3], [3,2],$ and $[2,1,2]$.
\begin{table}[h]
	\resizebox{\linewidth}{!}{
\begin{tabular}{cccc}
\hlinewd{1.5pt}
Utterance  & Speaker & MELD & EmotionLines \\
\hlinewd{0.8pt}

I'’m so sorry!    & Chandler     & sadness    & sadness   \\
Look!    & Chandler     & surprise   & surprise   \\
This guy fell asleep! & Chandler     & \textcolor{maroon}{anger}     & non-neutral      \\
% He fell asleep too!  & Chandler     & \textcolor{maroon}{anger}     & non-neutral \\ 
\hlinewd{0.8pt}
\end{tabular}
}
\caption{Difference in annotation between EmotionLines and MELD. %\textit{Non-neutral} signifies the case where annotators agreed that the emotion is not neutral but they could not reach consensus regarding the correct emotion label.
}
\label{tab:emovsmeld}
%\vspace{-0.2cm}
\end{table}

\begin{table*}[htbp!]
	\small
	\centering
	\begin{tabular}{ccccccc}
		%\cline{2-7}
		\hlinewd{1.5pt}
		\multicolumn{3}{c}{\text{}} & \multicolumn{2}{c}{{Incorrect Splits}} & \multicolumn{2}{c}{{Corrected Splits}} \\
		\cline{2-7}
		{Utterance}  & {Season} & {Episode} & Start Time   & End Time    & Start Time   & End Time    \\
  \hlinewd{0.8pt}

		Chris says they're closing & \multirow{2}{*}{3}    & \multirow{2}{*}{6}     & \multirow{2}{*}{00:05:57,023}  & \multirow{2}{*}{00:05:59,691}  & \multirow{2}{*}{00:05:57,023}  & \multirow{2}{*}{00:05:58,734} \\ 
		down the bar. &&&&&& \\
		No way!    & 3    & 6     & 00:05:57,023  & 00:05:59,691  & 00:05:58,734  & 00:05:59,691  \\
		\hlinewd{0.8pt}   
	\end{tabular}
	\caption{Example of timestamp alignment using the Gentle alignment tool.}
	\label{tab:pruning}
\end{table*}

\subsection{Dataset Exploration}

As mentioned before, we use seven emotions for the annotation, i.e., \textit{anger, disgust, fear, joy, neutral, sadness}, and \textit{surprise}, across the training, development, and testing splits (see Table~\ref{tab:emomeldvsemotionlines}). It can be seen that the emotion distribution in the dataset is expectedly non-uniform with the majority emotion being \emph{neutral}. 
\begin{table*}[h]
	\small
	\resizebox{\linewidth}{!}{
	\begin{tabular}{ccccccccc}
		\hlinewd{1.5pt}
		Utterance                                                       & Speaker     & Emotion & D\_ID & U\_ID & Season & Episode & StartTime  & EndTime   \\
    \hlinewd{0.8pt}
		But then who? The waitress I went out & \multirow{2}{*}{Joey}  & \multirow{2}{*}{surprise}  & \multirow{2}{*}{1} & \multirow{2}{*}{0}       & \multirow{2}{*}{9}   & \multirow{2}{*}{23}   & \multirow{2}{*}{00:36:40,364} & \multirow{2}{*}{00:36:42,824}\\ 
		with last month?   &&&&&&&&\\
				You know? Forget it!                    & Rachel & sadness   & 1 & 1       & 9   & 23   & 00:36:44,368 & 00:36:46,578 \\
		\hlinewd{0.8pt}
	\end{tabular}
}
	\caption{MELD dataset format for a dialogue. Notations: D\_ID = dialogue ID, U\_ID = utterance ID. StartTime and EndTime are in hh:mm:ss,ms format. %Note: A full length conversation is provided in the supplementary.
	}
	\label{tab:dataformat}
%	\vspace{-0.2cm}
\end{table*}
We have also converted these fine-grained emotion labels into more coarse-grained sentiment classes by considering \emph{anger, disgust, fear, sadness} as \emph{negative}, \emph{joy} as \emph{positive}, and \emph{neutral} as \emph{neutral} sentiment-bearing class. \emph{Surprise} is an example of a complex emotion which can be expressed with both positive and negative sentiment. The three annotators who performed the utterance annotation further annotated the \textit{surprise} utterances into either positive or negative sentiment classes. The entire sentiment annotation task reaches a Fleiss' kappa score of 0.91. The distribution of \emph{positive, negative, neutral} sentiment classes is given in Table~\ref{tab:emomeldvsemotionlines}.

% \begin{table}[t]
% 	\centering
% 	\scalebox{0.9}{
% 	\begin{tabular}{crrr}
% 		\hlinewd{1.5pt}
% 		& \multicolumn{3}{c}{{No. of Utterances}} \\
% 		\cline{2-4}
% 		{Sentiment Category} & Train      & Dev     & Test     \\
% 		\hline
% 		\hline
% 		negative         & 2945      & 406     & 833      \\
% 		neutral          & 4710      & 470     & 1256     \\
% 		positive         & 2334      & 233     & 521  \\
% 		\hlinewd{0.8pt}
% 	\end{tabular}}
% 	\caption{Coarse sentiment distribution in the dataset.}
% 	\label{tab:emodist2}
% \end{table}

\begin{table}[b]
	\centering
	\small
  \resizebox{\linewidth}{!}{
	\begin{tabular}{cccc}
		\hlinewd{1.5pt}
		{MELD Statistics} & {Train} & {Dev} & {Test} \\
    \hlinewd{0.8pt}
		{\# of modalities}           & \{a,v,t\}    & \{a,v,t\}    & \{a,v,t\}\\
		{\# of unique words}   & 10,643  & 2,384    & 4,361  \\
		{Avg./Max utterance length}    & 8.0/69   & 7.9/37  & 8.2/45  \\
		%{Max. utterance length}    & 69    & 37   & 45   \\
		{\# of dialogues}     & 1039   & 114  & 280   \\
		{\# of dialogues dyadic MELD} &2560 &270 &577\\
		{\# of utterances}    & 9989  & 1109  & 2610  \\
		{\# of speakers}     & 260   & 47   & 100   \\
		{Avg. \# of utterances per dialogue} & 9.6 & 9.7 & 9.3 \\
		{Avg. \# of emotions per dialogue} & 3.3   & 3.3  & 3.2  \\
		{Avg./Max \# of speakers per dialogue} &2.7/9 &3.0/8 & 2.6/8 \\
		%{Max \# of speakers per dialogue} &9 &8 & 8 \\
		{\# of emotion shift}     & 4003   & 427  & 1003  \\
		{Avg. duration of an utterance}   & 3.59s    & 3.59s  & 3.58s \\
		\hlinewd{0.8pt} 
	\end{tabular}
	}
	\caption{Dataset Statistics. \{a,v,t\} = \{audio, visual, text\}}
	\label{tab:datasetstat}
\end{table}

Table~\ref{tab:datasetstat} presents several key statistics of the dataset. The average utterance length -- i.e. number of words in an utterance -- is nearly the same across training, development, and testing splits. On average, three emotions are present in each dialogue of the dataset. The average duration of an utterance is $3.59$ seconds. The emotion shift of a speaker in a dialogue makes emotion recognition task very challenging. \begin{figure*}[htbp]
  \centering
	\includegraphics[width=0.81\linewidth]{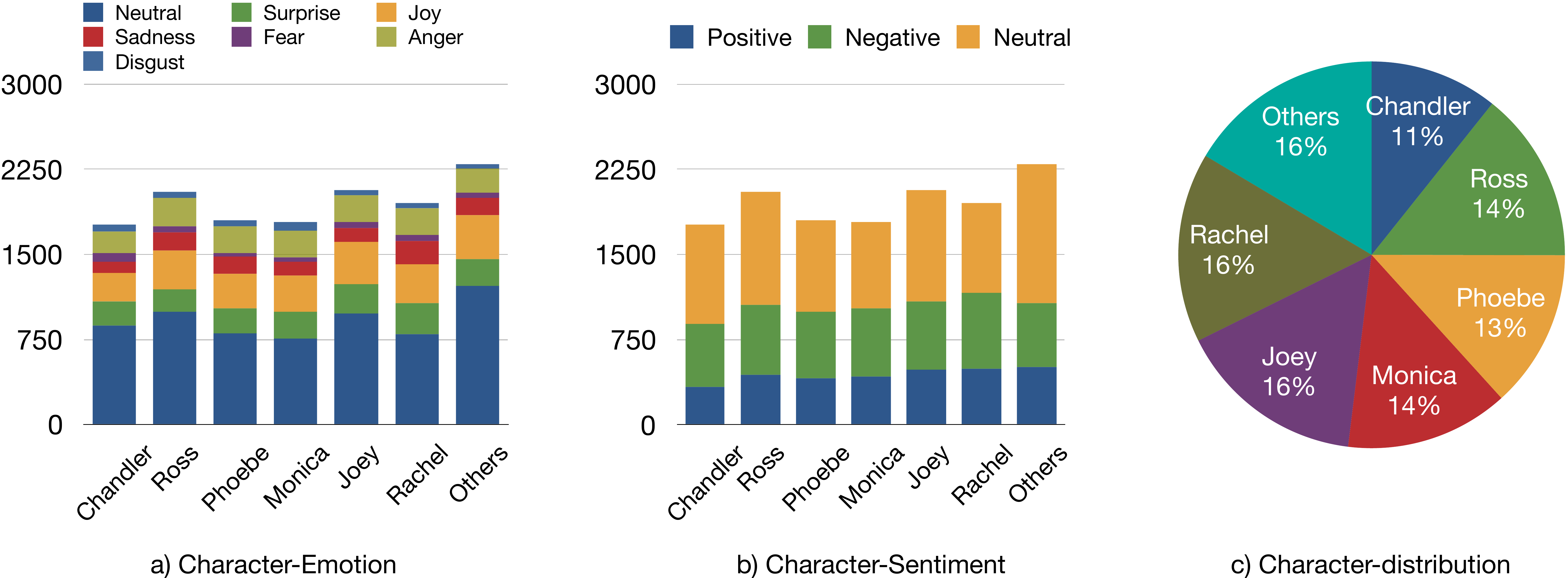}
	\caption{Character distribution across MELD.}
	\label{fig:character_dist}
\end{figure*}
We observe that the number of such emotion shifts in successive utterances of a speaker in a dialogue is very frequent: $4003$, $427$, and $1003$ in train/dev/test splits, respectively. Figure~\ref{fig:emoc} shows an example where speaker's emotion changes with time in the dialogue.

\paragraph{Character Distribution.}
In Figure~\ref{fig:character_dist}, we present the distributional details of the primary characters in MELD. Figure a and b illustrate the distribution across the emotion and sentiment labels, respectively. Figure c shows the overall coverage of the speakers across the dataset. Multiple infrequent speakers ($<1\%$ utterances) are grouped as \textit{Others}.

\subsection{Related Datasets}
\label{sec:rel}
Most of the available datasets in multimodal sentiment analysis and emotion recognition are non-conversational. MOSI~\cite{zadeh2016multimodal}, MOSEI~\cite{zadeh2018multimodal}, and MOUD~\cite{perez2013utterance} are such examples that have drawn significant interest from the research community. On the other hand, IEMOCAP and SEMAINE are two popular dyadic conversational datasets where each utterance in a dialogue is labeled by emotion.

\textbf{The SEMAINE Database%\footnote{\scriptsize{\url{https://sspnet.eu/avec2012/}}}
} 
is an audiovisual database created for building agents that can engage a person in a sustained and emotional conversation~\cite{mckeown2012semaine}. It consists of interactions involving a \textit{human} and an \textit{operator} (either a machine or a person simulating a machine). The dataset contains 150 participants, 959 conversations, each lasting around 5 minutes. A subset of this dataset was used in AVEC 2012's \textit{fully continuous sub-challenge}~\cite{schuller2012avec} that requires predictions of four continuous affective dimensions: \textit{arousal, expectancy, power,} and \textit{valence}. The gold annotations are available for every $0.2$ second in each video for a total of $95$ videos comprising $5,816$ utterances. 

% For utterance-level applications, annotations are typically approximated by taking an average of the continuous labels within the spoken utterance.

\textbf{The Interactive Emotional Dyadic Motion Capture Database (IEMOCAP)%\footnote{\scriptsize{\url{https://sail.usc.edu/iemocap/}}}
} consists of videos of dyadic conversations among pairs of 10 speakers spanning 10 hours of various dialogue scenarios~\cite{iemocap}. Videos are segmented into utterances with annotations of fine-grained emotion categories: \textit{anger, happiness, sadness, neutral, excitement,} and \textit{frustration}. IEMOCAP also provides continuous attributes: \textit{activation}, \textit{valence}, and \textit{dominance}. These two types of discrete and continuous emotional descriptors facilitate the complementary insights about the emotional expressions of humans and emotional communications between people. The labels in IEMOCAP were annotated by at least three annotators per utterance and self-assessment manikins (SAMs) were also employed to evaluate the corpus~\cite{bradley1994measuring}.

\subsection{Comparison with MELD}
\label{sec:comparison}
\begin{table}[b]
\centering
%\small
	\resizebox{\linewidth}{!}{
		\begin{tabular}{c|c|c|c|c|c|c|c}
			\hlinewd{1.5pt}
			\multirow{2}{*}{Dataset}&Type&\multicolumn{3}{c|}{$\#$ dialogues}&\multicolumn{3}{c}{$\#$ utterances}\\
			&&train&dev&test&train&dev&test\\
			\hlinewd{0.8pt}
			IEMOCAP&acted&\multicolumn{2}{c|}{120}&31&\multicolumn{2}{c|}{5810}&1623\\
			SEMAINE&acted&\multicolumn{2}{c|}{58}&22&\multicolumn{2}{c|}{4386}&1430\\
% 			\hline
% 			EmotionLines&720&80&200&10561&1178&2764\\
			MELD&acted&1039&114&280& 9989& 1109& 2610\\
			\hlinewd{.8pt} 
		\end{tabular}
		}
	\caption{Comparison among IEMOCAP, SEMAINE, and proposed MELD datasets}
	\label{tab:compare}
\end{table}
Both resources mentioned above are extensively used in this field of research and contain settings that are aligned to the components of MELD. However, MELD is different  in terms of both complexity and quantity. 
%\begin{figure}[b] 
%	\centering 
%	\includegraphics[width=\linewidth]{chart-meld}
%	\caption[]{Comparison between the distribution of common emotions between training splits of IEMOCAP and MELD.}
%	\label{fig:emotion_dist}
%\end{figure}
Both IEMOCAP and SEMAINE contain dyadic conversations, wherein the dialogues in MELD are multi-party. 
Multi-party conversations are more challenging compared to dyadic. They provide a flexible setting where multiple speakers can engage. From a research perspective, such availability also demands proposed dialogue models to be scalable towards multiple speakers.
MELD also includes more than 13000 emotion labeled utterances, which is nearly double  the annotated utterances in IEMOCAP and SEMAINE.  Table~\ref{tab:compare} provides information on the number of available dialogues and their constituent utterances for all three datasets, i.e., IEMOCAP, SEMAINE, and MELD. %As seen in the table, MELD contains the largest amount of dialogues (and utterances), which is significantly larger than the other two. MELD also has the most utterances as \textit{neutral} which emulates real-life conversations where the prevailing emotion is generally \textit{neutral}. 
\cref{tab:IEMOCAPvsMELD} shows the distribution for common emotions as well as highlights a few key statistics of IEMOCAP and MELD.  

\begin{table*}[h!]
\centering
%\small
	\resizebox{\linewidth}{!}{
		\begin{tabular}{c|c|c|c|c|c|c|c|c|c}
			\hlinewd{1.5pt}
			\multirow{3}{*}{Dataset}&\multicolumn{6}{c|}{ Emotions}&\multicolumn{3}{c}{Other Statistics}\\
			\cline{2-10}
			&{Happy/Joy} & {Anger} & {Disgust} &{Sadness} &{Surprise} &{Neutral}&\begin{tabular}{cc}&Avg.\\ &utterence length\end{tabular}&\begin{tabular}{cc}&$\#$Unique\\& words\end{tabular}&\begin{tabular}{cc}&Avg.\\ &conversation length\end{tabular}\\
			\hlinewd{0.8pt}
            IEMOCAP&648&1103&2&1084&107&1708&15.8&3,598&49.2\\
        MELD&2308&1607&361&1002&1636&6436&8.0&10,643&9.6\\
			\hlinewd{.8pt} 
		\end{tabular}
		}
	\caption{Comparison among IEMOCAP and proposed MELD datasets.}
	\label{tab:IEMOCAPvsMELD}
\end{table*}

\section{Experiments}
\label{sec:experiment}
% \begin{table}[t]
% 	\centering
% 	\large
% 	\resizebox{\linewidth}{!}{
% 	\begin{tabular}{c|cccccccc}
% 		\hlinewd{1.5pt}
% 		\multirow{2}{*}{Modality}&\multicolumn{8}{c}{{Emotions}}\\\cline{2-9}
% 		& ang & disg & fear & joy & neu & sad & surp & w-avg.\\\hline\hline
% 		text-CNN & 34.4 & 8.2 & 3.7 & 49.3 & 74.8 & 21.0 & 45.4 & 55.0 \\
% 		text &40.5 &2.0 &8.9 &50.2 &75.7 &24.1 &49.3 &57.0 \\
% 		audio & 35.1 &5.1 &5.5 &13.1 &65.5 &14.0 &20.4 &41.7 \\
% 		text+audio &43.3 &23.6 &9.3 &54.4 &76.6 &24.3 &51.0 &59.2 \\
% 		\hlinewd{0.8pt}
% 	\end{tabular}
% 	}
% 	\caption{Test-set F-score results of DialogueRNN for emotion classification in MELD. Note: \textit{w-avg} denotes weighted-average. text-CNN: CNN applied on text, contextual information were not used.}
% 	\label{tab:emotion_results_7way}
% \end{table}

\subsection{Feature Extraction}
\label{sec:feature_extraction}

%In our experiments, we provide multimodal baselines concerning two modalities: text and audio. 
%Features are extracted in two stages -- first, unimodal features are extracted from text and audio. Next, these unimodal features are fed to contextual-LSTM~\citep{porcon} for contextual unimodal feature representation. 
We follow \citet{porcon} to extract features for each utterance in MELD. For textual features, we initialize each token with pre-trained 300-dimensional GloVe vectors~\cite{pennington2014glove} and feed them to a 1D-CNN to extract 100 dimensional textual features. For audio, we use the popular toolkit openSMILE~\cite{eyben2010opensmile}, which extracts 6373 dimensional features constituting several low-level descriptors and various statistical functionals of varied vocal and prosodic features. %OpenSMILE extracts 6373 dimensional features using the ``IS13-ComParE'' configuration file, which constitutes of several low-level descriptors and various statistical functionals. % from audio segments. 
As the audio representation is high dimensional, we employ L2-based feature selection with sparse estimators, such as SVMs, to get a dense representation of the overall audio segment. For the baselines, we do not use visual features, as video-based speaker identification and localization is  an open problem. Bimodal features are obtained by concatenating audio and textual features.

\subsection{Baseline Models}
\label{sec:baseline_models}

To provide strong benchmarks for MELD, we perform experiments with multiple baselines. Hyperparameter details for each baseline %used in this paper 
can be found at \url{http://github.com/senticnet/meld}.

\textbf{text-CNN} applies CNN to the input utterances without considering the context of the conversation~\cite{kim2014convolutional}. This model represents the simplest baseline which does not leverage context or multimodality in its approach.

\textbf{bcLSTM} is a strong baseline proposed by~\citet{porcon}, which represents context using a bi-directional RNN. It follows a two-step hierarchical process that models uni-modal context first and then bi-modal context features. For unimodal text, a CNN-LSTM model extracts contextual representations for each utterance taking the GloVe embeddings as input. For unimodal audio, an LSTM model gets audio representations for each audio utterance feature vector. Finally, the contextual representations from the unimodal variants are supplied to the bimodal model for classification. bcLSTM does not distinguish among different speakers and models a conversation as a single sequence.

\textbf{DialogueRNN} represents the current state of the art for conversational emotion detection~\cite{majumder2018dialoguernn}. It is a strong baseline with effective mechanisms to model context by tracking individual speaker states throughout the conversation for emotion classification. DialogueRNN is capable of handling multi-party conversation so it can be directly applied on MELD. It employs three stages of gated recurrent units (GRU)~\cite{DBLP:journals/corr/ChungGCB14} to model emotional context in conversations. The spoken utterances are fed into two GRUs: \textit{global} and \textit{party GRU} to update the context and speaker state, respectively. In each turn, the party GRU updates its state based on 1) the utterance spoken, 2) the speaker's previous state, and 3) the conversational context summarized by the global GRU through an attention mechanism. Finally, the updated speaker state is fed into the \textit{emotion GRU} which models the emotional information for classification. Attention mechanism is used on top of the \textit{emotion GRU} to leverage contextual utterances by different speakers at various distances. To analyze the role of multimodal signals, we analyze DialogueRNN and bcLSTM on MELD for both uni and multimodal settings.
Training involved usage of class weights to alleviate imbalance issues.

\subsection{Results}
We provide results for the two tasks of sentiment and emotion classification on MELD. Table~\ref{tab:sentiment_results} shows the performance of sentiment classification by using  DialogueRNN, whose multimodal variant achieves the best performance (67.56\% F-score) surpassing multimodal bcLSTM (66.68\% F-score). Multimodal DialogueRNN also outperforms its unimodal counterparts. However, the improvement due to  fusion is about 1.4\% higher than the textual modality which suggests the possibility of further improvement through better fusion mechanisms. The textual modality outperforms the audio modality by about 17\%, which indicates the importance of spoken language in sentiment analysis. For positive sentiment, audio modality performs poorly. It would be interesting to analyze the clues specific to positive sentiment bearing utterances in MELD that the audio modality could not capture. Future work should aim for enhanced audio feature extraction schemes to improve the classification performance.
\begin{table*}[t!]
	\centering
	%\small
	\scalebox{0.85}{
	\begin{tabular}{cc|cccccccc}
		\hlinewd{1.5pt}
		\multicolumn{2}{c|}{\multirow{2}{*}{Models}}&\multicolumn{8}{c}{{Emotions}}\\\cline{3-10}
		&& anger & disgust & fear & joy & neutral & sadness & surprise& w-avg.\\
		\hlinewd{0.8pt}
		\multicolumn{2}{c|}{text-CNN} & 34.49 & 8.22 & 3.74 & 49.39 & 74.88 & 21.05 & 45.45 & 55.02 \\ \cline{1-2}
		\multirow{1}{*}{cMKL}&text+audio &39.50 &16.10 &3.75 &51.39 &72.73 &23.95 &46.25 & 55.51 \\ \cline{1-2}
		\multirow{3}{*}{bcLSTM}&text &42.06 &21.69 &7.75 &54.31 &71.63 &26.92 &48.15 & 56.44 \\
		&audio & 25.85 &6.06 &2.90 &15.74 &61.86 &14.71 &19.34 &39.08 \\
		&text+audio &43.39 &23.66 &9.38 &54.48 &76.67 &24.34 &51.04 &59.25 \\ \cline{1-2}
		\multirow{3}{*}{DialogueRNN}&text &40.59 &2.04 &8.93 &50.27 &75.75 &24.19 &49.38 &57.03 \\
		&audio & 35.18 &5.13 &5.56 &13.17 &65.57 &14.01 &20.47 &41.79 \\
		&text+audio &43.65 &7.89 &11.68 &54.40 &77.44 &34.59 &52.51 &\textbf{60.25} \\
		\hlinewd{0.8pt}
	\end{tabular}
	}
	\caption{Test-set weighted F-score results of DialogueRNN for emotion classification in MELD. Note: \textit{w-avg} denotes weighted-average. text-CNN and cMKL: contextual information were not used.}
	\label{tab:emotion_results_7way}
	%\vspace{-0.2cm}
\end{table*}
Table~\ref{tab:emotion_results_7way} presents the results of the baseline models on MELD emotion classification. The performance on the emotion classes \textit{disgust}, \textit{fear}, and \textit{sadness} are particularly poor. The primary reason for this is the inherent imbalance in the dataset which has fewer training instances for these mentioned emotion classes (see Table~\ref{tab:emomeldvsemotionlines}). We partially tackle this by using class-weights as hyper-parameters. 

Yet, the imbalance calls for further improvement for future work to address. We also observe high mis-classification rate between the \textit{anger}, \textit{disgust}, and \textit{fear} emotion categories as these emotions have subtle differences among them causing harder disambiguation. Similar to sentiment classification trends, the textual classifier outperforms (57.03\% F-score) the audio classifier (41.79\% F-score). 

Multimodal fusion helps in improving the emotion recognition performance by 3\%. However, multimodal classifier performs worse than the textual classifier in classifying sadness. To analyze further, we also run experiments on 5-class emotions by dropping the infrequent \textit{fear} and \textit{disgust} emotions (see Table~\ref{tab:emotion_results_5way}). Not surprisingly, the results improve over the 7-class setting with significantly better performance by the multimodal variant.

Overall, emotion classification performs poorer than sentiment classification. This observation is expected as emotion classification deals with classification with more fine-grained classes. 
\begin{table}[b]
	\centering
	\small
	\resizebox{\linewidth}{!}{
	\begin{tabular}{cc|cccccccc}
		\hlinewd{1.5pt}
		&\multirow{2}{*}{Mode}&\multicolumn{8}{c}{{Emotions}}\\\cline{3-10}
		&& ang & joy & neu & sad & surp & w-avg.\\\hlinewd{0.8pt}
		bcLSTM&T+A &45.9 &52.2 &77.9 &11.2 &49.9 &60.6 \\ \cline{1-2}
		\multirow{3}{*}{dRNN$^*$}&T &41.7 &53.7 &77.8 &21.2 &47.7 & 60.8 \\
		&A & 34.1 &18.8 &66.2 &16.0 &16.6 &44.3 \\
		&T+A &48.2 &53.2 &77.7 &20.3 &48.5 &\textbf{61.6} \\
		\hlinewd{0.8pt}
		\multicolumn{8}{l}{$^*$dRNN: DialogueRNN, T: text, A: audio }
	\end{tabular}
	}
	\caption{Test-set weighted F-score results of DialogueRNN for 5-class emotion classification in MELD. Note: \textit{w-avg} denotes weighted-average. \textit{surp}: surprise emotion.}
	\label{tab:emotion_results_5way}
\end{table}

\subsection{Additional Analysis}
\label{sec:analysis}

\paragraph{Role of Context.}
One of the main purposes of MELD is to train contextual modeling in a conversation for emotion recognition. Table~\ref{tab:emotion_results_7way} and~\ref{tab:sentiment_results} show that the improvement over the non-contextual model such as text-CNN -- which only uses a CNN (see Section~\ref{sec:feature_extraction}) -- is 1.4\% to 2.5\%.
\paragraph{Inter-speaker influence.} One of the important considerations while modeling conversational emotion dynamics is the influence of fellow speakers in the multi-party setting. We analyze this factor by looking at the activation of the attention module on the  \textit{global GRU} in DialogueRNN. We observe that in $63\%$ ($882/1381$) of the correct test predictions, the highest historical attention is given to utterances from different speakers. This significant proportion suggests inter-speaker influence to be an important parameter. 
\begin{table}[htbp!]
	\centering
	\small
	\resizebox{\linewidth}{!}{
	%\scalebox{0.9}{
	\begin{tabular}{cc|cccc}
		\hlinewd{1.5pt}
		&\multirow{2}{*}{Mode}&\multicolumn{4}{c}{{Sentiments}}\\\cline{3-6}
		&& pos. & neg. & neu. &w-avg.\\
		\hlinewd{0.8pt}
		&text-CNN & 53.23 & 55.42 & 74.69 & 64.25 \\ \cline{1-2}
		bcLSTM&T+A &74.68 &57.87 &60.04 &66.68 \\ \cline{1-2}
		\multirow{3}{*}{dRNN$^*$}&T & 54.35 & 60.10 & 74.94 & 66.10 \\
		&A & 25.47 & 45.53 & 62.33 & 49.61 \\ 
		&T+A & 54.29 & 58.18 & 78.40 & \textbf{67.56}\\
		\hlinewd{0.8pt}
	\end{tabular}
	}
	\caption{Test set weighted F-score results of DialogueRNN for sentiment classification in MELD. %Note: \textit{w-avg}: weighted-average. 
	}
	\label{tab:sentiment_results}
	%\vspace{-0.3cm}
\end{table}
Unlike DialogueRNN, bcLSTM does not utilize speaker information while detecting emotion. Table ~\ref{tab:emotion_results_7way} shows that in all the experiments, DialogueRNN outperforms bcLSTM by 1-2\% margin. This result supports the claim by~\citet{majumder2018dialoguernn} that speaker-specific modeling of emotion recognition is beneficial as it helps in improving context representation and incorporates important clues such as inter-speaker relations.

% How many shifts have been detected correctly
\paragraph{Emotion shifts.} The ability to anticipate the emotion shifts within speakers throughout the course of a dialogue has synergy with better emotion classification. In our results, DialogueRNN achieves a recall of $66\%$ for detecting emotion shifts. %In other words, DialogueRNN can predict a shift in emotion $66\%$ of times. 
However, in the ideal scenario, we would want to detect shift along with the correct emotion class. For this setting, DialogueRNN gets a recall of $36.7\%$. The deterioration observed is expected as solving both tasks together has a higher complexity. Future methods would need to improve upon their capabilities of detecting shifts to improve the emotion classification.
% Also attention to past and future

\begin{figure}[htbp!] 
 	\centering 
	\includegraphics[width=\linewidth]{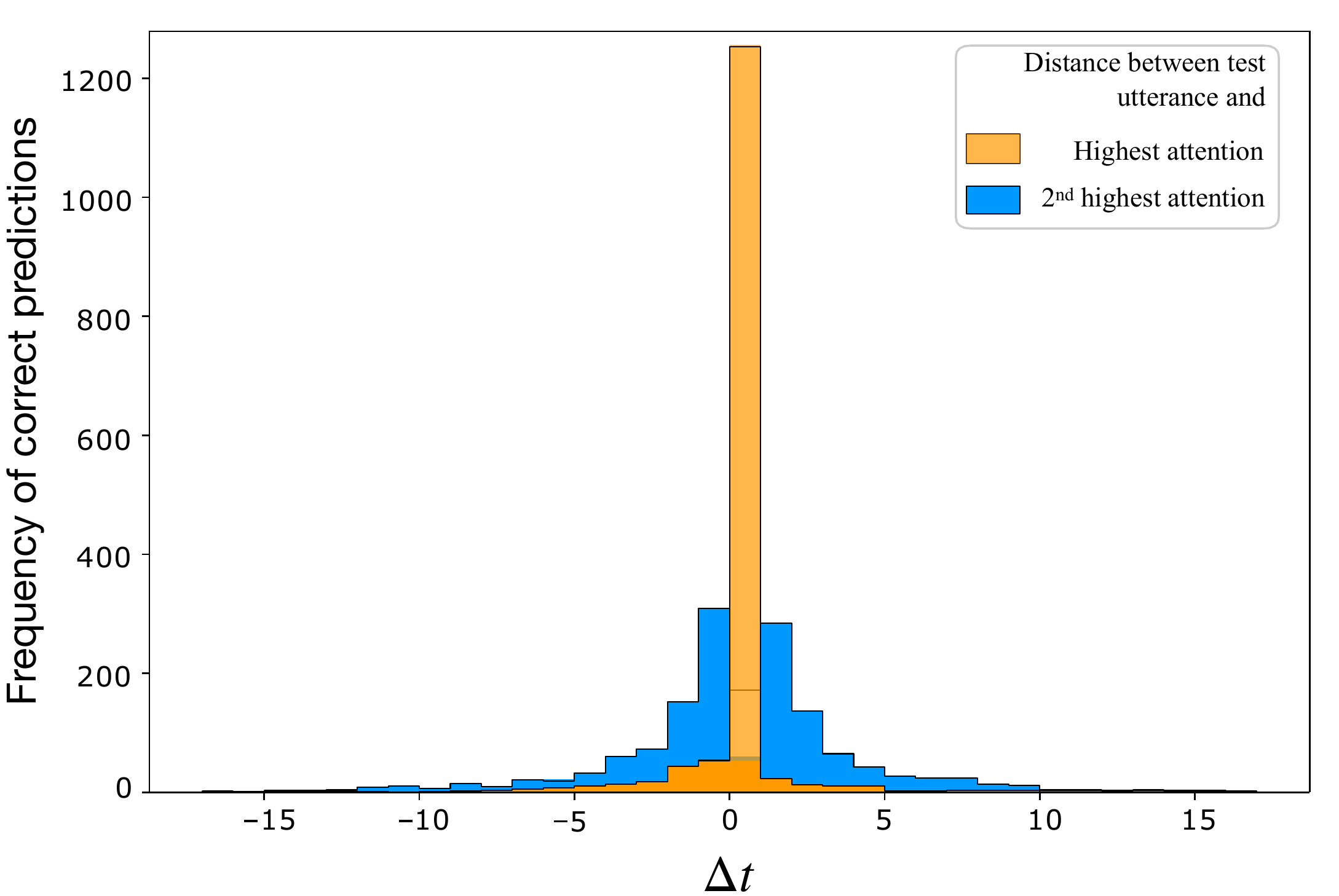}
	\caption[]{Histogram of $\Delta t=$ distance between the target and its context utterance based on \textit{emotion GRU} attention scores.}
	\label{fig:context_dist}
	%\vspace{-0.2cm}
\end{figure}

\paragraph{Contextual distance.}
Figure~\ref{fig:context_dist} presents the distribution of distances between the target utterance and its second highest attended utterance within the conversation by DialogueRNN in its \textit{emotion GRU}. For the highest attention, the model largely focuses on utterances nearby to the target utterance. However, the dependency on distant utterances increases with the second highest attention. Moreover, it is interesting to see that the dependency exists both towards the historical and the future utterances, thus incentivizing utilization of bi-directional models.
\section{Future Directions}
\label{sec:future}

Future research using this dataset should focus on improving contextual modeling. Helping models reason about their decisions, exploring emotional influences, and identifying emotion shifts are  promising aspects.
Another direction is to use visual information available in the raw videos. Identifying face of the speaker in a video where multiple other persons are present is very challenging. 
This is the case for MELD too as it is a multi-party dataset. Enhancements can be made by extracting relevant visual features through processes utilizing audio-visual speaker diarization. Such procedures would enable utilizing a visual modality in the baselines. In our results, audio features do not help significantly. Thus, we believe that it is necessary to improve the feature extraction for these auxiliary modalities in order to improve the performance further.

So far, we have only used concatenation as a feature fusion approach, and showed that it outperforms the  unimodal baselines by about 1-3\%. We believe there is room for further improvement using other more advanced fusion methods such as MARN~\citep{zadeh2018multimodal}.

\section{Applications of MELD}
\label{sec:appli}

MELD has multiple use-cases. It can be used to train emotion classifiers to be further used as emotional receptors in generative dialogue systems. These systems can be used to generate empathetic responses~\cite{zhou2017emotional}. It can also be used for emotion and personality modeling of users in conversations~\citep{li2016persona}. 

By being multimodal, MELD can also be used  to train multimodal dialogue systems. Although by itself it is not large enough to train an end-to-end dialogue system (Table~\ref{tab:data}), the procedures used to create MELD can be adopted to generate a large-scale corpus from any multimodal source such as popular sitcoms. We define \emph{multimodal dialogue system} as a platform where the system has access to the speaker's voice and facial expressions which it exploits to generate responses. Multimodal dialogue systems can be very useful for real time personal assistants such as Siri, Google Assistant where the users can use both voice and text and facial expressions to communicate.

\section{Conclusion}
\label{sec:conclusion}

In this work, we introduced MELD, a multimodal multi-party conversational emotion recognition dataset. We described the process of building this dataset,  and provided results obtained with strong baseline methods applied on this dataset. MELD  contains raw videos, audio segments, and transcripts for multimodal processing. Additionally, we also provide the features used in our baseline experiments. We believe this dataset will also be useful as a training corpus for both conversational emotion recognition and multimodal empathetic response generation. %Overall, %MELD has a strong potential to help conversation understanding research. 
%Designing multimodal fusion algorithm and developing ERC framework are the two most important future works that can be done using this dataset. 
Building upon this dataset, future research can explore the design of efficient multimodal fusion algorithms, novel ERC frameworks, as well as the  extraction of new features from the audio, visual, and textual modalities. %The MELD dataset can be freely downloaded from \url{https://affective-meld.github.io/}.
%Future works should also focus on the rich feature extraction from audio, visual and textual modalities.

% \section*{Acknowledgments}
\section*{Acknowledgments}
This material is based in part upon work supported by the National Science Foundation (grant \#1815291), by the John Templeton Foundation (grant \#61156), and by DARPA (grant \#HR001117S0026-AIDA-FP-045). 

\bibliography{acl2019}

\begin{thebibliography}{22}
\expandafter\ifx\csname natexlab\endcsname\relax\def\natexlab#1{#1}\fi

\bibitem[{Bradley and Lang(1994)}]{bradley1994measuring}
Margaret~M Bradley and Peter~J Lang. 1994.
\newblock Measuring emotion: the self-assessment manikin and the semantic
  differential.
\newblock \emph{Journal of behavior therapy and experimental psychiatry},
  25(1):49--59.

\bibitem[{Busso et~al.(2008)Busso, Bulut, Lee, Kazemzadeh, Mower, Kim, Chang,
  Lee, and Narayanan}]{iemocap}
Carlos Busso, Murtaza Bulut, Chi-Chun Lee, Abe Kazemzadeh, Emily Mower, Samuel
  Kim, Jeannette~N Chang, Sungbok Lee, and Shrikanth~S Narayanan. 2008.
\newblock Iemocap: Interactive emotional dyadic motion capture database.
\newblock \emph{Language resources and evaluation}, 42(4):335--359.

\bibitem[{Chen et~al.(2018)Chen, Hsu, Kuo, Ku et~al.}]{chen2018emotionlines}
Sheng-Yeh Chen, Chao-Chun Hsu, Chuan-Chun Kuo, Lun-Wei Ku, et~al. 2018.
\newblock Emotionlines: An emotion corpus of multi-party conversations.
\newblock \emph{arXiv preprint arXiv:1802.08379}.

\bibitem[{Chung et~al.(2014)Chung, G{\"{u}}l{\c{c}}ehre, Cho, and
  Bengio}]{DBLP:journals/corr/ChungGCB14}
Junyoung Chung, {\c{C}}aglar G{\"{u}}l{\c{c}}ehre, KyungHyun Cho, and Yoshua
  Bengio. 2014.
\newblock \href {http://arxiv.org/abs/1412.3555} {{Empirical Evaluation of
  Gated Recurrent Neural Networks on Sequence Modeling}}.
\newblock \emph{CoRR}, abs/1412.3555.

\bibitem[{Eyben et~al.(2010)Eyben, W{\"o}llmer, and
  Schuller}]{eyben2010opensmile}
Florian Eyben, Martin W{\"o}llmer, and Bj{\"o}rn Schuller. 2010.
\newblock Opensmile: the munich versatile and fast open-source audio feature
  extractor.
\newblock In \emph{Proceedings of the 18th ACM international conference on
  Multimedia}, pages 1459--1462. ACM.

\bibitem[{Hazarika et~al.(2018)Hazarika, Poria, Zadeh, Cambria, Morency, and
  Zimmermann}]{hazarika2018conversational}
Devamanyu Hazarika, Soujanya Poria, Amir Zadeh, Erik Cambria, Louis-Philippe
  Morency, and Roger Zimmermann. 2018.
\newblock Conversational memory network for emotion recognition in dyadic
  dialogue videos.
\newblock In \emph{{NAACL}}, volume~1, pages 2122--2132.

\bibitem[{Kim(2014)}]{kim2014convolutional}
Yoon Kim. 2014.
\newblock Convolutional neural networks for sentence classification.
\newblock In \emph{{EMNLP}}, pages 1746--1751.

\bibitem[{Li et~al.(2016)Li, Galley, Brockett, Spithourakis, Gao, and
  Dolan}]{li2016persona}
Jiwei Li, Michel Galley, Chris Brockett, Georgios Spithourakis, Jianfeng Gao,
  and Bill Dolan. 2016.
\newblock A persona-based neural conversation model.
\newblock In \emph{{ACL}}, volume~1, pages 994--1003.

\bibitem[{Majumder et~al.(2019)Majumder, Poria, Hazarika, Mihalcea, Gelbukh,
  and Cambria}]{majumder2018dialoguernn}
Navonil Majumder, Soujanya Poria, Devamanyu Hazarika, Rada Mihalcea, Alexander
  Gelbukh, and Erik Cambria. 2019.
\newblock {DialogueRNN}: An attentive {RNN} for emotion detection in
  conversations.
\newblock \emph{Thirty-Third AAAI Conference on Artificial Intelligence}.

\bibitem[{McKeown et~al.(2012)McKeown, Valstar, Cowie, Pantic, and
  Schroder}]{mckeown2012semaine}
Gary McKeown, Michel Valstar, Roddy Cowie, Maja Pantic, and Marc Schroder.
  2012.
\newblock The semaine database: Annotated multimodal records of emotionally
  colored conversations between a person and a limited agent.
\newblock \emph{Affective Computing, IEEE Transactions on}, 3(1):5--17.

\bibitem[{Pennington et~al.(2014)Pennington, Socher, and
  Manning}]{pennington2014glove}
Jeffrey Pennington, Richard Socher, and Christopher Manning. 2014.
\newblock Glove: Global vectors for word representation.
\newblock In \emph{EMNLP}, pages 1532--1543.

\bibitem[{P{\'e}rez-Rosas et~al.(2013)P{\'e}rez-Rosas, Mihalcea, and
  Morency}]{perez2013utterance}
Ver{\'o}nica P{\'e}rez-Rosas, Rada Mihalcea, and Louis-Philippe Morency. 2013.
\newblock Utterance-level multimodal sentiment analysis.
\newblock In \emph{ACL (1)}, pages 973--982.

\bibitem[{Poria et~al.(2017)Poria, Cambria, Hazarika, Mazumder, Zadeh, and
  Morency}]{porcon}
Soujanya Poria, Erik Cambria, Devamanyu Hazarika, Navonil Mazumder, Amir Zadeh,
  and Louis-Philippe Morency. 2017.
\newblock Context-dependent sentiment analysis in user-generated videos.
\newblock In \emph{{ACL}}, pages 873--883.

\bibitem[{Poria et~al.(2019)Poria, Majumder, Mihalcea, and
  Hovy}]{poria2019emotion}
Soujanya Poria, Navonil Majumder, Rada Mihalcea, and Eduard Hovy. 2019.
\newblock Emotion recognition in conversation: Research challenges, datasets,
  and recent advances.
\newblock \emph{arXiv preprint arXiv:1905.02947}.

\bibitem[{Rashkin et~al.(2018)Rashkin, Smith, Li, and
  Boureau}]{rashkin2018know}
Hannah Rashkin, Eric~Michael Smith, Margaret Li, and Y-Lan Boureau. 2018.
\newblock I know the feeling: Learning to converse with empathy.
\newblock \emph{arXiv preprint arXiv:1811.00207}.

\bibitem[{Schuller et~al.(2012)Schuller, Valster, Eyben, Cowie, and
  Pantic}]{schuller2012avec}
Bj{\"o}rn Schuller, Michel Valster, Florian Eyben, Roddy Cowie, and Maja
  Pantic. 2012.
\newblock Avec 2012: the continuous audio/visual emotion challenge.
\newblock In \emph{Proceedings of the 14th ACM international conference on
  Multimodal interaction}, pages 449--456. ACM.

\bibitem[{Serban et~al.(2017)Serban, Sordoni, Lowe, Charlin, Pineau, Courville,
  and Bengio}]{serban2017hierarchical}
Iulian~Vlad Serban, Alessandro Sordoni, Ryan Lowe, Laurent Charlin, Joelle
  Pineau, Aaron~C Courville, and Yoshua Bengio. 2017.
\newblock A hierarchical latent variable encoder-decoder model for generating
  dialogues.
\newblock In \emph{AAAI}, pages 3295--3301.

\bibitem[{Wollmer et~al.(2013)Wollmer, Weninger, Knaup, Schuller, Sun, Sagae,
  and Morency}]{wollmer2013youtube}
Martin Wollmer, Felix Weninger, Timo Knaup, Bjorn Schuller, Congkai Sun, Kenji
  Sagae, and Louis-Philippe Morency. 2013.
\newblock Youtube movie reviews: Sentiment analysis in an audio-visual context.
\newblock \emph{IEEE Intelligent Systems}, 28(3):46--53.

\bibitem[{Zadeh et~al.(2016{\natexlab{a}})Zadeh, Baltru{\v{s}}aitis, and
  Morency}]{zadeh2016deep}
Amir Zadeh, Tadas Baltru{\v{s}}aitis, and Louis-Philippe Morency.
  2016{\natexlab{a}}.
\newblock Deep constrained local models for facial landmark detection.
\newblock \emph{arXiv preprint arXiv:1611.08657}.

\bibitem[{Zadeh et~al.(2018)Zadeh, Liang, Poria, Cambria, and
  Morency}]{zadeh2018multimodal}
Amir Zadeh, Paul~Pu Liang, Soujanya Poria, Erik Cambria, and Louis-Philippe
  Morency. 2018.
\newblock Multimodal language analysis in the wild: Cmu-mosei dataset and
  interpretable dynamic fusion graph.
\newblock In \emph{{ACL}}, volume~1, pages 2236--2246.

\bibitem[{Zadeh et~al.(2016{\natexlab{b}})Zadeh, Zellers, Pincus, and
  Morency}]{zadeh2016multimodal}
Amir Zadeh, Rowan Zellers, Eli Pincus, and Louis-Philippe Morency.
  2016{\natexlab{b}}.
\newblock Multimodal sentiment intensity analysis in videos: Facial gestures
  and verbal messages.
\newblock \emph{IEEE Intelligent Systems}, 31(6):82--88.

\bibitem[{Zhou et~al.(2017)Zhou, Huang, Zhang, Zhu, and
  Liu}]{zhou2017emotional}
Hao Zhou, Minlie Huang, Tianyang Zhang, Xiaoyan Zhu, and Bing Liu. 2017.
\newblock Emotional chatting machine: Emotional conversation generation with
  internal and external memory.
\newblock \emph{arXiv preprint arXiv:1704.01074}.

\end{thebibliography}
\bibliographystyle{acl_natbib}

\end{document}